\pdfoutput=1
\documentclass[11pt]{article}

\usepackage[preprint]{acl}

\usepackage{times}
\usepackage{latexsym}

\usepackage[T1]{fontenc}
\usepackage[utf8]{inputenc}

\usepackage{microtype}

\usepackage{inconsolata}

\usepackage{graphicx}

\usepackage{booktabs}
\usepackage{amsmath}
\PassOptionsToPackage{table,xcdraw}{xcolor}
\usepackage{xcolor}
\usepackage{multirow}
\usepackage{amsfonts}
\usepackage{enumitem}

\usepackage{colortbl}
\usepackage[normalem]{ulem}
\useunder{\uline}{\ul}{}

\usepackage{subcaption}

\title{D-CoDe: Scaling Image-Pretrained VLMs to Video via\\Dynamic Compression and Question Decomposition}

\author{
\textbf{Yiyang Huang}, 
\textbf{Yizhou Wang}, 
\textbf{Yun Fu} \\
Northeastern University \\
huang.yiyan@northeastern.edu, yunfu@ece.neu.edu
}

\begin{document}
\maketitle

\begin{abstract}
Video large language models (Vid-LLMs), which excel in diverse video-language tasks, can be effectively constructed by adapting image-pretrained vision-language models (VLMs). However, this adaptation remains challenging, as it requires processing dense and temporally extended visual inputs that exceed the capacity of image-based models.  
This paper identifies the perception bottleneck and token overload as key challenges in extending image-based VLMs to the video domain. To address these issues, we propose D-CoDe, a training-free adaptation framework that incorporates dynamic compression and question decomposition. Specifically, dynamic compression alleviates the perception bottleneck through adaptive selection of representative frames and content-aware aggregation of spatial tokens, thereby reducing redundancy while preserving informative content. In parallel, question decomposition mitigates token overload by reformulating the original query into sub-questions, guiding the model to focus on distinct aspects of the video and enabling more comprehensive understanding. 
Experiments demonstrate that D-CoDe effectively improves video understanding across various benchmarks. Furthermore, strong performance on the challenging long-video benchmark highlights the potential of D-CoDe in handling complex video-language tasks. Code is available at \url{https://github.com/hukcc/D-CoDe}.
\end{abstract}
\begin{figure}[t] 
    \centering
    \includegraphics[width=0.925\linewidth]{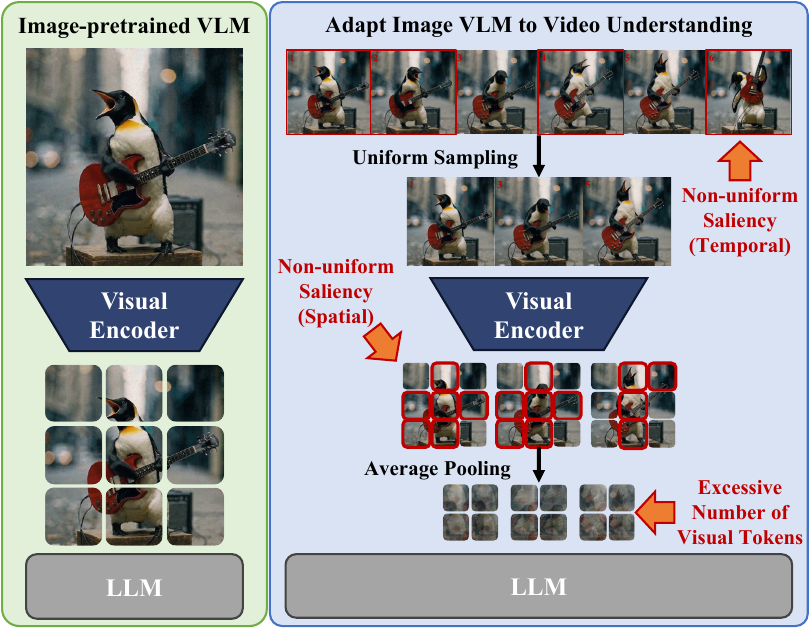}  
    \caption{Adapting image-pretrained VLMs to video faces two major challenges: the perception bottleneck, in which salient information is unevenly distributed across spatial and temporal dimensions, limiting the effectiveness of static compression in preserving key visual cues; and token overload, where video inputs yield substantially more visual tokens than images, exceeding the model's capacity for comprehensive understanding.}
    \label{fig:teasor}
\end{figure}
\section{Introduction}

Video large language models (Vid-LLMs) integrate video inputs with textual instructions and demonstrate strong performance across a wide range of video-language tasks. However, constructing Vid-LLMs directly from pre-trained large language models is constrained by the scarcity of high-quality video-text data~\cite{scarcity}. A more data-efficient alternative is to adapt image-pretrained vision-language models (VLMs), leveraging the structural similarity between images and videos.

Approaches for adapting image-pretrained VLMs to video can be broadly divided into training-required and training-free methods. Training-required methods typically fine-tune the visual encoder or cross-modal connector \cite{videochat, videochat2}, align visual features between images and videos \cite{videollava}, incorporate additional modalities to broaden task coverage \cite{videollama, videollama2}, or apply techniques like Direct Preference Optimization (DPO) \cite{llavanext-video} and slow-fast architectures \cite{lita} to enhance temporal modeling and factual consistency. Despite their effectiveness, these methods often incur high computational cost. In contrast, training-free methods leverage image-pretrained VLMs without additional tuning, yet still achieve competitive performance. Representative examples include IG-VLM~\cite{ig-vlm}, which constructs a grid-view image from sampled frames; FreeVA~\cite{freeva}, which performs frame-level temporal aggregation; SF-LLaVA~\cite{sf-llava}, which employs a slow-fast architecture; and TS-LLaVA~\cite{ts-llava}, which adopts a thumbnail-and-sampling strategy to generate compact and informative visual prompts.

However, despite their efficiency, training-free methods face two key challenges that limit scalability: perception bottleneck and token overload, as shown in Figure~\ref{fig:teasor}. Perception bottleneck arises from static compression strategies such as uniform frame sampling and spatial average pooling, which treat all content equally and discard salient information unevenly distributed across temporal and spatial dimensions, thereby limiting the model's ability to capture fine-grained visual cues. Token overload, in turn, occurs when compressed video inputs still contain substantially more visual tokens than static images, exceeding the processing capacity of image-pretrained VLMs and hindering the modeling of long-range dependencies and complex spatio-temporal structures essential for comprehensive understanding.

To overcome these challenges, we propose D-CoDe, a training-free framework that extends image-pretrained VLMs to video understanding by integrating dynamic compression and question decomposition. Specifically, dynamic compression augments temporal uniform sampling by selecting supplementary frames from segments exhibiting greater semantic variation, then filters out uninformative spatial tokens and merges semantically similar ones, thereby reducing redundancy while preserving informative visual cues. In parallel, question decomposition enhances the model's capacity to interpret dense visual inputs by reformulating complex queries into focused sub-questions, guiding attention to distinct aspects of the video and enabling comprehensive understanding.

Experiments show that D-CoDe consistently improves performance across a range of video understanding benchmarks, including multiple-choice VideoQA (NExT-QA, EgoSchema, IntentQA) and open-ended VideoQA (MSVD-QA, MSRVTT-QA, TGIF-QA, ANet-QA), which cover diverse video types from first- and third-person perspectives and span durations from short clips to long-form content. Notably, D-CoDe is the first training-free method to surpass training-required models on EgoSchema, a challenging benchmark involving long-form egocentric videos and schema-driven questions.

Our contributions are summarized as follows:
\begin{itemize}[nosep]
\item We analyze the key challenges in adapting image-pretrained VLMs to video understanding, focusing on the perception bottleneck and token overload.
\item We introduce D-CoDe, a training-free framework that addresses the perception bottleneck via content-aware dynamic compression and mitigates token overload through question decomposition.
\item Extensive experiments across various benchmarks validate the effectiveness of D-CoDe. In particular, strong performance on the long-video task highlights its potential for complex video-language understanding.
\end{itemize}

\section{Related Work}

\subsection{Training-based Video-LLMs}
Training-based Video-LLMs learn video understanding through fine-tuning image-pretrained VLMs on large-scale video datasets. Video-ChatGPT~\cite{vid-chatgpt} extends LLaVA~\cite{llava} with temporal and spatial pooling. 
% \textcolor{blue}{
VidF4~\cite{VidF4} enhances BLIP-2 with frame scoring for adaptive sampling. LongVU~\cite{LongUV} applies query-guided frame selection and token merging across frames to compress videos.
% } 
VideoChat~\cite{videochat} employs Q-Former for token compression, and VideoChat2~\cite{videochat2} improves alignment and instruction tuning. Video-LLaVA~\cite{videollava} introduces a shared projector to unify image and video encoders. Video-LLaMA~\cite{videollama, videollama2} integrates video, audio, and language for multimodal tasks. LLaVA-NeXT-Video~\cite{llavanext-video} fine-tunes LLaVA-NeXT~\cite{llavanext} with DPO~\cite{dpo} to improve performance. LITA~\cite{lita} adopts a slow-fast architecture~\cite{slowfastnetwork, audiovisualslowfast} for spatio-temporal modeling. While effective, these methods are computationally expensive.

\subsection{Training-free Video-LLMs}
Training-free Video-LLMs, in contrast, extend image-pretrained VLMs to video without additional fine-tuning. IG-VLM~\cite{ig-vlm} constructs a grid-view image from video frames and feeds it into a frozen LLM with prompt-based adaptation. FreeVA~\cite{freeva} explores temporal aggregation but relies on few frames. SF-LLaVA~\cite{sf-llava} applies a slow-fast design inspired by action recognition~\cite{slowfastnetwork, audiovisualslowfast, lita}. TS-LLaVA~\cite{ts-llava} uses a thumbnail-and-sampling strategy to create compact visual prompts with supplementary tokens. While promising, few training-free approaches directly address the perception bottleneck caused by fixed compression strategies and the token overload resulting from the limited visual token capacity of image-pretrained VLMs, leaving a gap that this work aims to fill.

\section{Method}
\subsection{Challenges in Image-to-Video Adaptation}
Given that video inputs yield substantially more visual tokens than static images, effective token compression and comprehensive understanding of the resulting representations are crucial for adapting image-pretrained VLMs to video. However, the perception bottleneck hinders efficient compression with minimal information loss, while token overload limits comprehensive interpretation of the compressed tokens, as their number still exceeds the capacity of image-pretrained VLMs.

\begin{figure}[t]
    \centering

    \begin{subfigure}{0.925\linewidth}
        \centering
        \includegraphics[width=\linewidth]{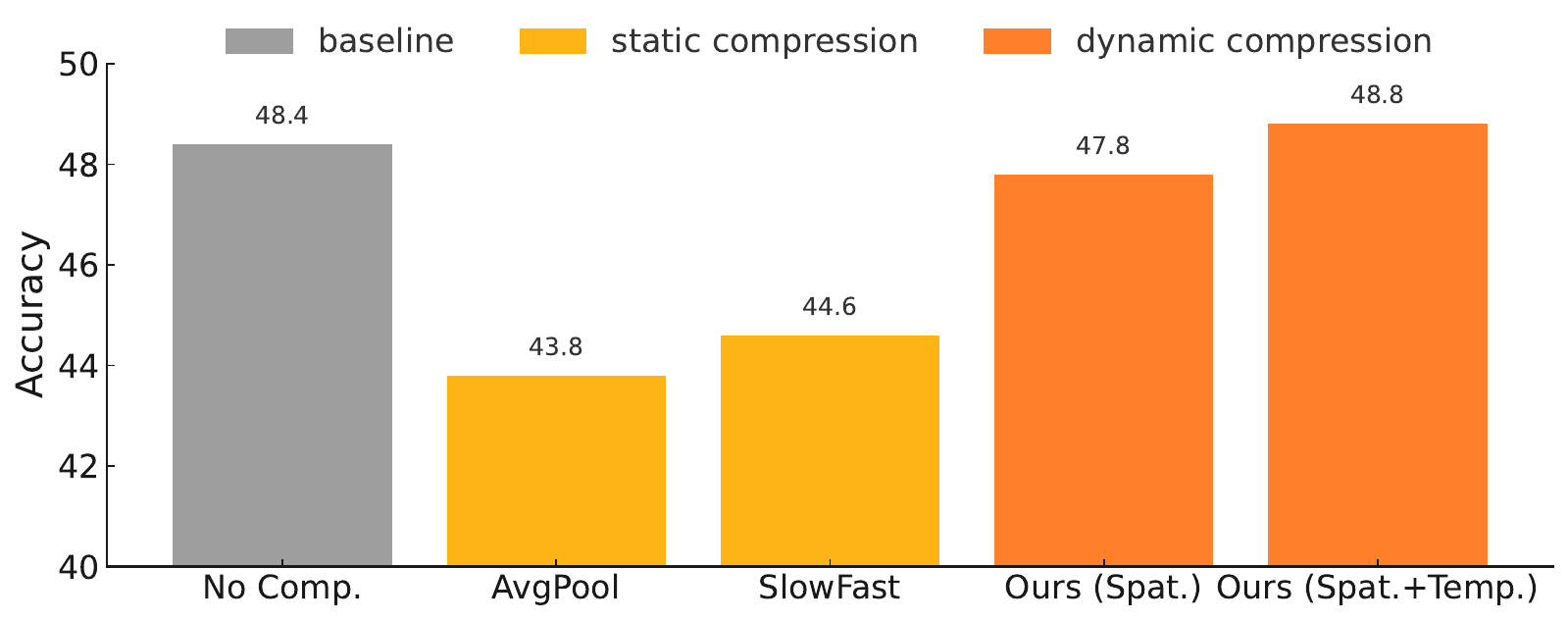}
        \caption{Accuracy comparison on EgoSchema with 5 input frames. The X-axis denotes compression strategies; the Y-axis indicates accuracy.}
        \label{fig:problem-vis-sub1}
    \end{subfigure}

    \vspace{1pt}

    \begin{subfigure}{0.925\linewidth}
        \centering
        \includegraphics[width=\linewidth]{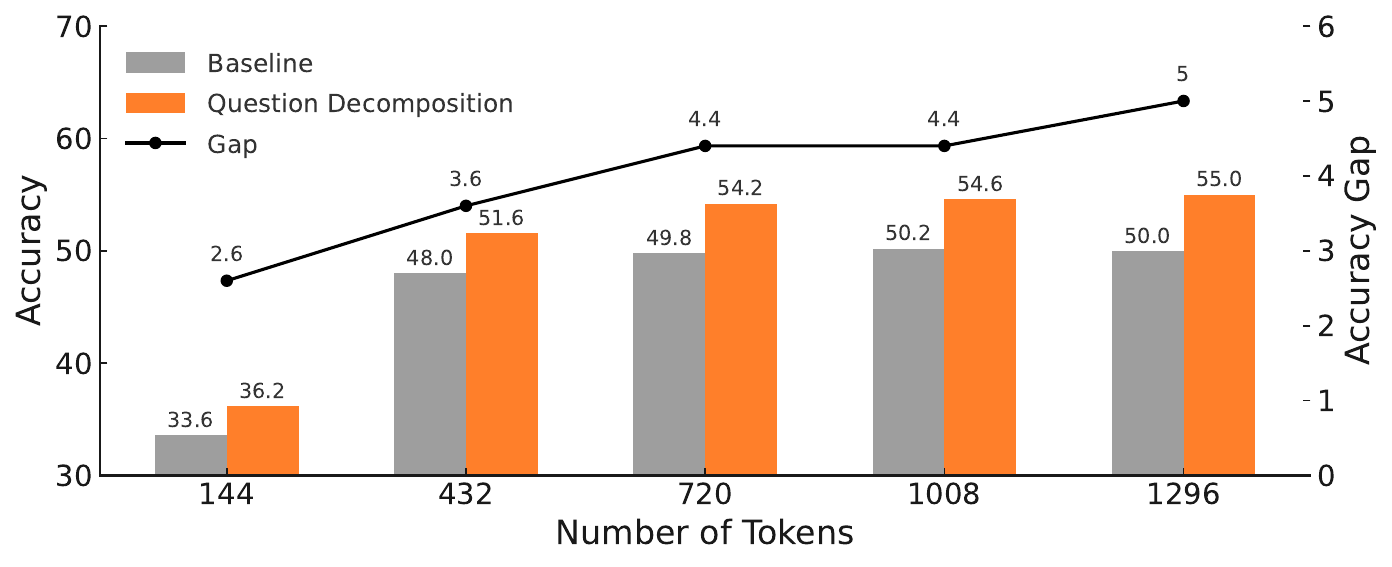}
        \caption{Accuracy of the baseline and its question decomposition variant on EgoSchema using 10 input frames, under varying visual token counts determined by different top-$k$ activation retention ratios. The X-axis indicates the number of input tokens, and the Y-axis indicates accuracy.}
        \label{fig:problem-vis-sub2}
    \end{subfigure}

    \caption{(a) Static compression treats all content uniformly, discarding informative cues that are dynamically distributed across temporal and spatial dimensions, thereby limiting fine-grained perception. In contrast, dynamic compression better preserves key visual cues across both dimensions. (b) As the number of input tokens increases, the accuracy of the baseline saturates, indicating limited utility of excessive tokens. In contrast, question decomposition consistently expands the accuracy gap, demonstrating its ability to more effectively leverage large token inputs.}
    \label{fig:problem-vis}
\end{figure}

\subsubsection{Perception Bottleneck}
\label{sec:perception_bottleneck}
The perception bottleneck arises from static compression strategies such as uniform frame sampling and spatial average pooling, which, despite their efficiency, lack semantic adaptivity and tend to discard informative cues that are unevenly distributed across temporal and spatial dimensions. This limits the model's ability to capture fine-grained visual details. Figure~\ref{fig:problem-vis-sub1} illustrates this issue by comparing the performance of the 7B LLaVA-NeXT model on the EgoSchema benchmark using 5 input frames under different compression strategies. Compared to the uncompressed baseline, static methods lead to notable performance degradation. In contrast, our dynamic compression alleviates this drop and even surpasses the baseline by reducing redundancy while preserving informative visual cues.

\subsubsection{Token Overload}
\label{sec:token_overload}
Token overload arises when video inputs, even after compression, contain substantially more visual tokens than static images, exceeding the processing capacity of image-pretrained VLMs. As a result, performance no longer improves with increasing token count, as the model cannot effectively interpret the excess information. Figure~\ref{fig:problem-vis-sub2} illustrates this effect by comparing the performance of the vanilla 7B LLaVA-NeXT and its question decomposition variant on EgoSchema using 10 input frames, under varying numbers of visual tokens determined by different top-$k$ activation retention ratios. The vanilla model exhibits initial improvements followed by a performance plateau as the token count increases, indicating a typical token overload effect. In contrast, question decomposition consistently outperforms the baseline, with a widening accuracy gap that demonstrates superior scalability to larger token volumes.

\begin{figure*}[ht]
    \centering
    \includegraphics[width=0.925\textwidth]{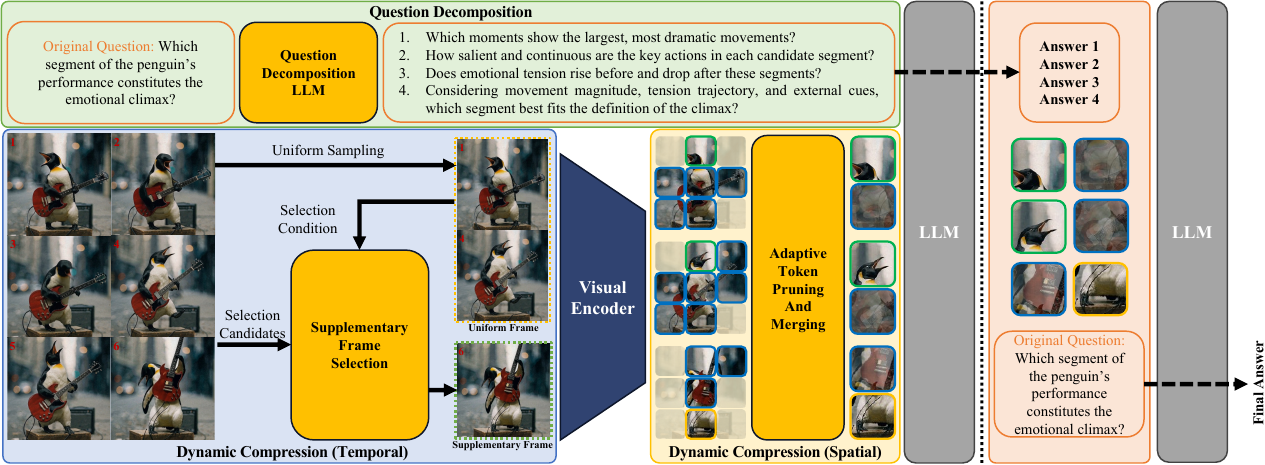}
    \caption{The D-CoDe pipeline consists of two components: dynamic compression and question decomposition. Dynamic compression augments temporal uniform sampling by selecting supplementary frames to retain informative segments, then discards uninformative spatial tokens and merges semantically similar ones to reduce redundancy while preserving essential visual information. Question decomposition reformulates complex queries into sub-questions, guiding the model to attend to diverse aspects of the video and enabling comprehensive understanding.}
    \label{fig:pipeline}
\end{figure*}

\subsection{D-CoDe: Dynamic Compression and Question Decomposition Adapting}
Our observations suggest that the perception bottleneck and token overload hinder effective compression and interpretation of visual inputs, thus limiting the adaptation of image-pretrained VLMs to video understanding. To tackle these challenges, we introduce D-CoDe, a training-free framework that scales image-pretrained VLMs to video by integrating dynamic compression and question decomposition, as shown in Figure~\ref{fig:pipeline}. The dynamic compression module augments uniform sampling with supplementary frames to retain informative segments, then filters out uninformative spatial tokens and merges semantically similar ones to balance semantic fidelity and token efficiency. The question decomposition module reformulates complex queries into focused sub-questions, guiding the model's attention to distinct aspects of the video and enabling comprehensive understanding.

\subsubsection{Formulation of Video-LLM Inference}
Let $\mathcal{V} = \{ I_t \}_{t=1}^{T}$ denote a video consisting of $T$ frames. For each frame $I_t$, visual features are extracted using a pretrained image encoder (e.g., CLIP~\cite{clip}), yielding:
\begin{equation}
\mathbf{F}_t = \text{VisualEnc}(I_t).
\end{equation}

The resulting sequence of visual features is denoted as $\mathbf{F}_{1:T} = \{\mathbf{F}_1, \mathbf{F}_2, \dots, \mathbf{F}_T\}$. This visual representation, together with the query $Q$, is then provided to the LLM to generate the final answer:
\begin{equation}
A_{\text{final}} = \text{LLM}(\mathbf{F}_{1:T}, Q).
\end{equation}

\begin{figure}[t] 
    \centering
    \includegraphics[width=0.925\linewidth]{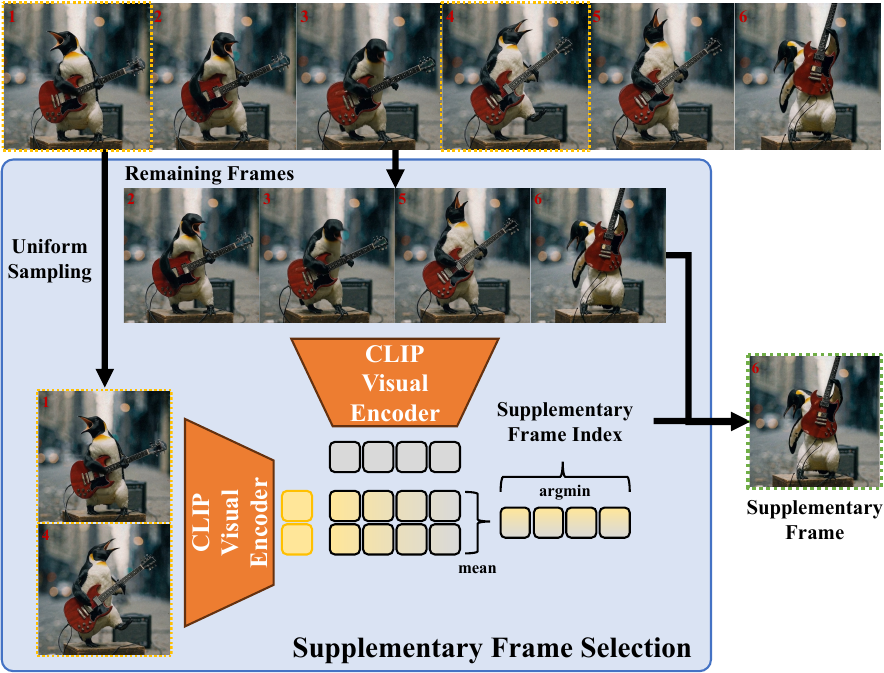}  
    \caption{To mitigate the temporal perception bottleneck, that is, to avoid missing informative video content, supplementary frames are selected based on their semantic dissimilarity to uniformly sampled ones, where similarity is measured using global features extracted by the CLIP visual encoder.}
    \label{fig:Supplementary_Frame_Selection}
\end{figure}

\subsubsection{Dynamic Compression}
As discussed in Section~\ref{sec:perception_bottleneck}, the perception bottleneck stems from static compression strategies that fail to retain salient information unevenly distributed across temporal and spatial dimensions. To address this, we introduce dynamic compression, which integrates semantic-aware frame selection with adaptive spatial token pruning and merging, thereby enhancing fine-grained detail retention while reducing redundancy.

As illustrated in Figure~\ref{fig:Supplementary_Frame_Selection}, to adapt temporal granularity based on content complexity, a two-stage strategy is employed to select $N$ representative frames from a video $\mathcal{V}$ of $T$ frames. In the first stage, $\lfloor \alpha \cdot N \rfloor$\footnotemark frames are uniformly sampled across the sequence, where $\alpha \in (0,1)$ denotes the uniform sampling ratio, forming $\mathcal{V}_{\text{uniform}}$ that provides coarse temporal coverage. To emphasize informative segments, the remaining $(N - \lfloor \alpha \cdot N \rfloor)$ frames are iteratively selected from the unchosen candidates. At each iteration, the frame with the lowest average semantic similarity to the current selected set $\mathcal{V}_{\text{selected}}$ is identified as the supplementary frame $I^{*}$ and appended to $\mathcal{V}_{\text{selected}}$:
\begin{equation}
I^* = \mathop{\arg\min}_{I_m \in \mathcal{V} \setminus \mathcal{V}_{\text{selected}}} \frac{1}{|\mathcal{V}_{\text{selected}}|} \sum_{I_n \in \mathcal{V}_{\text{selected}}} s_{m,n}.
\end{equation}
Here, $\mathcal{V}_{\text{selected}}$ is initialized with $\mathcal{V}_{\text{uniform}}$, and the similarity $s_{m,n}$ between frames $I_m$ and $I_n$ is computed as the cosine similarity between their CLIP-based global features, $\mathbf{g}_t = \operatorname{CLIP_v}(I_t)$:
\begin{equation}
s_{t,t'} = \frac{\langle \mathbf{g}_t , \mathbf{g}_{t'} \rangle}{\|\mathbf{g}_{t}\|_2 \cdot \|\mathbf{g}_{t'}\|_2}, \quad \forall t,t' \in \{1,\dots,T\}.
\end{equation}
The selection process continues until $N$ frames are chosen, forming a frame set that balances temporal coverage and highlights informative segments.

\footnotetext{$\lfloor \cdot \rfloor$ denotes the floor function, which rounds a value down to the nearest integer.}

\begin{figure}[t] 
    \centering
    \includegraphics[width=0.925\linewidth]{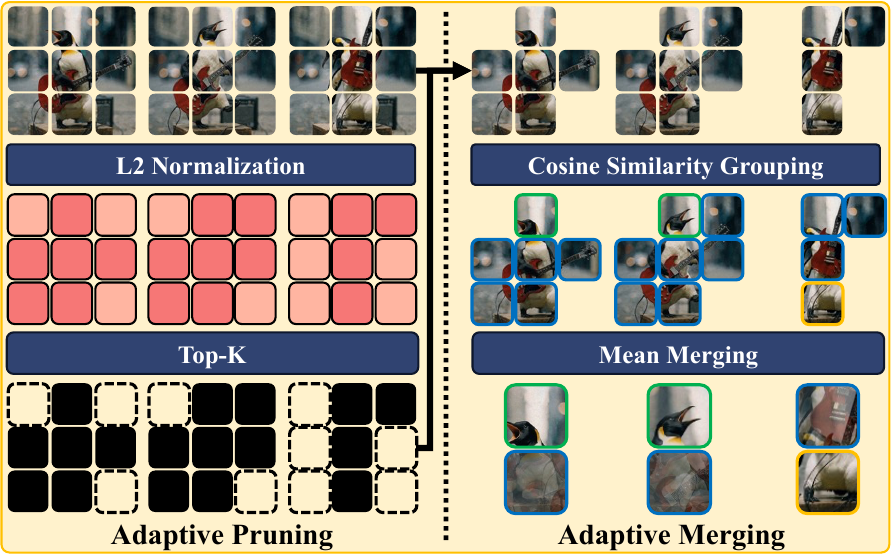}  
    \caption{To mitigate the spatial perception bottleneck, spatial tokens are first pruned based on their $\ell_2$ activation magnitudes. The remaining informative tokens are then grouped according to cosine similarity and aggregated via mean pooling, thereby reducing redundancy while preserving semantic fidelity.}
    \label{fig:adaptive_puning_merging}
\end{figure}

Furthermore, to reduce spatial redundancy while preserving semantic information, token compression is applied to each selected frame, which involves pruning uninformative visual tokens based on their activation magnitudes and merging tokens according to their cosine similarity, as illustrated in Figure~\ref{fig:adaptive_puning_merging}. Specifically, given a set of $M$ visual tokens $\mathbf{F} = \{ \mathbf{f}_i \}_{i=1}^{M}$ extracted from a selected frame, the $\ell_2$ norm of each token is computed as a proxy for salience, indicating its relative contribution to the overall visual representation:
\begin{equation}
a_i = \|\mathbf{f}_i\|_2.
\end{equation}
The top-\( \lfloor \beta \cdot M \rfloor \) tokens exhibiting the highest activation magnitudes are retained, where \( \beta \in (0,1) \) specifies the retention ratio:
\begin{equation}
\mathbf{F}_{\text{selected}} = \left\{ \mathbf{f}_j \,\middle|\, j \in \operatorname{TopK}\left(\{ a_i \}_{i=1}^{M}, \lfloor \beta \cdot M \rfloor \right) \right\}.
\end{equation}
To further eliminate redundancy within $\mathbf{F}_{\text{selected}}$, a greedy token merging algorithm is applied. Let $\pi$ denote a permutation over the indices of $\mathbf{F}_{\text{selected}}$ such that the tokens are sorted in descending order of activation magnitudes, i.e., $a_{\pi(1)} \geq a_{\pi(2)} \geq \dots \geq a_{\pi(\lfloor \beta M \rfloor)}$. The unmerged tokens are then traversed iteratively in the order defined by $\pi$, with each token considered as a potential anchor for redundancy merging. For each anchor token $\mathbf{f}_{\pi(i)}$, its cosine similarity with other unmerged tokens is computed as:
\begin{equation}
\text{sim}(\mathbf{f}_{\pi(i)}, \mathbf{f}_j) = \frac{\langle \mathbf{f}_{\pi(i)}, \mathbf{f}_j \rangle}{\|\mathbf{f}_{\pi(i)}\|_2 \cdot \|\mathbf{f}_j\|_2}.
\end{equation}
Based on the computed similarities, tokens whose similarity to the anchor exceeds a predefined threshold $\tau \in (0, 1)$ are deemed redundant and grouped with the anchor token $\mathbf{f}_{\pi(i)}$ to form a redundancy cluster:
\begin{equation}
\mathcal{N}_{\pi(i)} = \left\{ j \neq \pi(i)\ \middle|
\begin{aligned}
&\text{sim}(\mathbf{f}_{\pi(i)}, \mathbf{f}_j) \geq \tau,\ \\
&\mathbf{f}_j\ \text{is unmerged}
\end{aligned}
\right\}.
\end{equation}
To consolidate the cluster, a representative token is obtained by averaging the anchor and its redundant counterparts:
\begin{equation}
\mathbf{f}_{\pi(i)}^{\text{rep}} = \frac{1}{1 + |\mathcal{N}_{\pi(i)}|} ( \mathbf{f}_{\pi(i)} + \sum_{j \in \mathcal{N}_{\pi(i)}} \mathbf{f}_j ).
\end{equation}
After computing the representative token $\mathbf{f}_{\pi(i)}^{\text{rep}}$, all tokens in the cluster $\mathcal{N}_{\pi(i)}$ are marked as inactive and skipped in subsequent iterations. This process repeats until all tokens are either merged or selected as anchors, resulting in a compressed token set $\mathbf{F}_{\text{compressed}} = \{\mathbf{f}_{\pi(i)}^{\text{rep}}\}$ for each selected frame.

To form the final visual representation, the compressed token sets from all $N$ selected frames are concatenated as follows:
\begin{equation}
\mathbf{F}_{\text{final}} = \operatorname{Concat}\left( \mathbf{F}_{\text{compressed}}^{(k)} \right)_{k=1}^{N}.
\end{equation}
The resulting compact sequence is subsequently fed into the LLM for downstream processing.

\begin{table}[t]
\centering
% \small
{\fontsize{8pt}{10pt}\selectfont
\caption{Prompt Template for Question Decomposition}
\begin{tabular}{p{7.5cm}}
\toprule
\textbf{Question Decomposition Prompt} \\ \midrule
I am working on a video understanding task. Your job is to break down the given question into a series of subquestions that guide the model toward solving the problem. The subquestions should focus on temporal and dynamic aspects of the video, rather than just static information that could be answered from a single frame. I will provide a question, and you should output the corresponding subquestions in English. \\
\\

Question: ``\{\textit{user question here}\}'' \\
\\

Output the subquestions as a Python list of strings. \\
\bottomrule
\end{tabular}
\label{tab:gpt3.5_prompt}
}
\end{table}
\subsubsection{Question Decomposition}
Although dynamic compression reduces redundancy and preserves essential visual information, it still produces a number of visual tokens that exceed the capacity of image-pretrained VLMs. As a result, the model fails to comprehensively interpret the compressed tokens, a limitation known as token overload (Section~\ref{sec:token_overload}). To address this, we introduce question decomposition, which reformulates the original query into a sequence of focused sub-questions, directing the model's attention to distinct aspects of the video and enabling progressive understanding.

To implement this strategy, we construct a structured system prompt (Table~\ref{tab:gpt3.5_prompt}) to guide the generation of focused sub-questions. Given this prompt, a sequence of sub-questions is derived from the input query $Q$ using a pretrained question decomposition LLM $\mathcal{M}$ with temperature $t$:
\begin{equation}
{Q_1, Q_2, \dots, Q_n} = \mathcal{M}(Q, t),
\end{equation}
where each $Q_i$ targets a distinct aspect of the video, such as character location, actions, interactions, or scene transitions.

Subsequently, each sub-question $Q_i$ is independently processed by the LLM, conditioned on the shared visual input $\mathbf{F}_{\text{final}}$:
\begin{equation}
A_i = \text{LLM}(\mathbf{F}_{\text{final}}, Q_i), \quad i = 1, 2, \dots, n.
\end{equation}

Finally, the responses $\{A_1, A_2, \dots, A_n\}$ are concatenated to form an auxiliary prompt segment. This aggregated textual input, together with the original query $Q$ and the compressed visual token sequence $\mathbf{F}_{\text{final}}$, is then fed into the LLM to generate the final answer:
\begin{equation}
A_{\text{final}} = \text{LLM}(\mathbf{F}_{\text{final}}, \texttt{Concat}(A_1, \dots, A_n), Q).
\end{equation}

\section{Experiments}

\subsection{Benchmarks and Metrics}
\textbf{Multiple Choice VideoQA Benchmarks.} The multiple choice setting tests model's ability to select the correct answer from a set of options. We evaluate on three benchmarks: NExT-QA~\cite{xiao2021next} for causal and temporal understanding, EgoSchema~\cite{mangalam2023egoschema} for schema-level interpretation in egocentric videos, and IntentQA~\cite{li2023intentqa} for intention recognition from subtle cues. Accuracy is used as the evaluation metric.

\textbf{Open-Ended VideoQA Benchmarks.} The open-ended setting requires generating natural language answers based on video content. We evaluate on four benchmarks: MSVD-QA~\cite{chen2011collecting}, based on textual descriptions of short clips; MSRVTT-QA~\cite{xu2016msr}, featuring diverse web videos with complex scenes; TGIF-QA~\cite{TGIF-QA}, focusing on repetition counting and state transitions in GIFs; and ActivityNet-QA~\cite{ActivityNet-QA} (abbreviated as ANet-QA), comprising long videos with rich activity semantics. Response quality is assessed using GPT-based metrics: GPT-Accuracy for factual correctness, and GPT-Score (0–5) for completeness and fluency. All evaluations use \texttt{gpt-3.5-turbo-0125} for consistency~\cite{wu2024freeva}.

\subsection{Implementation Details}
D-CoDe is built upon the image-pretrained LLaVA-NeXT~\cite{llavanext} with 7B parameters. Following SF-LLaVA~\cite{sf-llava}, we adopt rotary position embeddings (RoPE)~\cite{rope} with a scaling factor of 2 to extend the context length to 8192 tokens. For each input video, we sample $N$ frames, where $N$ is empirically determined based on the average video length of the corresponding dataset. All frames are uniformly resized to $336 \times 336$ to construct the visual input sequence. In the dynamic compression module, we use a uniform frame sampling ratio $\alpha = 0.85$, retain salient tokens with a ratio $\beta = 0.625$, and merge semantically similar tokens using a cosine similarity threshold $\tau = 0.9$. For question decomposition, we use \texttt{gpt-3.5-turbo-0125} as the question decomposition LLM, with a temperature of $t = 0.5$ to balance diversity and consistency. The number of generated sub-questions $n$ is not constrained. All experiments are conducted on a single NVIDIA RTX A6000 GPU.

\begin{table}[]
\centering
% \caption{Accuracy (\%) of multiple-choice VideoQA models on NExT-QA, EgoSchema, and IntentQA. The best result for each dataset is shown in bold, and the second-best is underlined.}
\caption{Results on Multiple Choice Benchmarks}
\label{tab:mc_results}
% {\fontsize{10pt}{12pt}\selectfont
% {\fontsize{9pt}{10.5pt}\selectfont
{\fontsize{8pt}{9.5pt}\selectfont

\begin{tabular}{cccc}
\toprule
                                            & \multicolumn{3}{c}{Multiple Choice VideoQA (Accuracy)} \\ \cline{2-4} 
\multirow{-2}{*}{Method}                    & NextQA           & EgoSchema        & IntentQA         \\ \midrule
\multicolumn{4}{c}{Training-Required Methods}                                                                                                     \\ \midrule
Video-LLaVA                                 & 60.5             & 37.0             & N/A              \\
Video-LLaMA2                                & N/A              & 51.7             & N/A              \\
MovieChat+                                  & 54.8             & {\ul 56.4}       & N/A              \\
Vista-LLaMA                                 & 60.7             & N/A              & N/A              \\ \midrule
\multicolumn{4}{c}{Training-Free Methods}                                                                                                         \\ \midrule
DeepStack-L                                 & 61.0             & 38.4             & N/A              \\
$M^3$                                       & 63.1             & 36.8             & 58.8             \\
IG-VLM                                      & 63.1             & 35.8             & 60.3             \\
SF-LLaVA                                    & 64.2             & 47.2             & 60.1             \\
TS-LLaVA                                    & {\ul 66.5}       & 50.2             & {\ul 61.7}       \\
\rowcolor[HTML]{C0C0C0} 
D-CoDe (Ours)                               & \textbf{68.3}    & \textbf{58.0}    & \textbf{64.2}    \\
\bottomrule
\end{tabular}
}
\end{table}

\subsection{Comparison on Multiple Choice VideoQA}
Table~\ref{tab:mc_results} summarizes the multiple choice VideoQA results. D-CoDe consistently outperforms all prior methods, including both training-free and training-required models. 

Notably, on EgoSchema, a challenging benchmark with long-form egocentric videos and schema-based questions where training-required models usually perform better, D-CoDe surpasses the second-best training-free method TS-LLaVA by 7.8\%. It is also the first training-free model to outperform all training-required ones, exceeding the best of them, MovieChat+, by 1.6\%. These results demonstrate the effectiveness of our question decomposition strategy for understanding complex video content.

\subsection{Comparison on Open-Ended VideoQA}
\label{sec:open-end-results}
Table~\ref{tab:openended_results} presents the open-ended VideoQA results. Compared to multiple-choice tasks, these benchmarks involve simpler questions, often asking what, who, or yes/no, which are less suitable for decomposition. Therefore, only the dynamic compression module is applied. Despite this, D-CoDe outperforms most existing methods, including both training-free and training-required models. 

D-CoDe achieves the highest accuracy on MSVD-QA (80.0\%) and TGIF-QA (79.1\%), which focus on visual recognition and temporal reasoning, respectively. These results demonstrate its effectiveness in preserving fine-grained visual details. D-CoDe also performs well on the more challenging ActivityNet-QA, which consists of long videos with complex activities, reaching 56.4\% accuracy and a GPT-Score of 3.4.

On MSRVTT-QA, D-CoDe performs slightly below SF-LLaVA and TS-LLaVA, likely due to the dataset's frequent scene transitions that favor models with slow-fast processing structures.

\begin{table}[]
\centering
\caption{Results on Open-Ended Benchmarks}
\label{tab:openended_results}
{\fontsize{8pt}{9.5pt}\selectfont
% \small
\begin{tabular}{ccccc}
\toprule
                         & \multicolumn{4}{c}{Open-Ended VideoQA (Accuracy/Score)}               \\ \cline{2-5} 
\multirow{-2}{*}{Method} & MSVD           & MSRVTT & TGIF           & ANet           \\ \midrule
\multicolumn{5}{c}{Training-Required Methods}                                                                                                                    \\ \midrule
Video-LLaMA              & 51.6\scalebox{0.7}{/2.5}          & 29.6\scalebox{0.7}{/1.8}  & N/A               & 12.4\scalebox{0.7}{/1.1}          \\
Video-LLaMA2             & 70.9\scalebox{0.7}{/3.8}          & N/A       & N/A            & 50.2\scalebox{0.7}{/3.3}          \\
Video-ChatGPT            & 64.9\scalebox{0.7}{/3.3}          & 49.3\scalebox{0.7}{/2.8}  & 51.4\scalebox{0.7}{/3.0}          & 35.2\scalebox{0.7}{/2.7}          \\
VideoGPT+                & 72.4\scalebox{0.7}{/3.9}          & 60.6\scalebox{0.7}{/3.6}  & 74.6\scalebox{0.7}{/4.1}          & 50.6\scalebox{0.7}{/3.6}          \\
Video-LLaVA              & 70.7\scalebox{0.7}{/3.9}          & 59.2\scalebox{0.7}{/3.5}  & 70.0\scalebox{0.7}{/4.0}          & 45.3\scalebox{0.7}{/3.3}          \\
MovieChat                & 75.2\scalebox{0.7}{/3.8}          & 52.7\scalebox{0.7}{/2.6}  & N/A               & 45.7\scalebox{0.7}{/3.4}          \\
MovieChat+               & 76.5\scalebox{0.7}{/3.9}          & 53.9\scalebox{0.7}{/2.7}  & N/A               & 48.1\scalebox{0.7}{/3.4}          \\
VideoChat                & 56.3\scalebox{0.7}{/2.8}          & 45.0\scalebox{0.7}{/2.5}  & 34.4\scalebox{0.7}{/2.3}          & 26.5\scalebox{0.7}{/2.2}          \\
VideoChat2               & 70.0\scalebox{0.7}{/3.9}          & 54.1\scalebox{0.7}{/3.3}  & N/A               & 49.1\scalebox{0.7}{/3.3}          \\
Vista-LLaMA              & 65.3\scalebox{0.7}{/3.6}          & 60.5\scalebox{0.7}{/3.3}  & N/A               & 48.3\scalebox{0.7}{/3.3}          \\
LLaMA-VID                & 69.7\scalebox{0.7}{/3.7}          & 57.7\scalebox{0.7}{/3.2}  & N/A               & 47.4\scalebox{0.7}{/3.3}          \\
PLLaVA                   & 76.6\scalebox{0.7}{/4.1}          & 62.0\scalebox{0.7}{/3.5}  & 77.5\scalebox{0.7}{/4.1}          & 56.3\scalebox{0.7}{/3.5}          \\ \midrule
\multicolumn{5}{c}{Training-Free Methods}                                                                                                                        \\ \midrule
FreeVA                   & 73.8\scalebox{0.7}{/4.1}          & 60.0\scalebox{0.7}{/3.5}  & N/A               & 51.2\scalebox{0.7}{/3.5}          \\
DeepStack-L              & 76.0\scalebox{0.7}{/4.0}          & N/A                        & N/A               & 49.3\scalebox{0.7}{/3.1}          \\
IG-VLM                   & 78.8\scalebox{0.7}{/4.1}          & 63.7\scalebox{0.7}{/3.5}  & 73.0\scalebox{0.7}{/4.0}          & 54.3\scalebox{0.7}{/3.4}          \\
SF-LLaVA                 & {\ul79.1}\scalebox{0.7}{/4.1}     & \textbf{65.8}\scalebox{0.7}{/3.6}  & {\ul78.7}\scalebox{0.7}{/4.2}          & 55.5\scalebox{0.7}{/3.4}          \\
TS-LLaVA                 & 79.0\scalebox{0.7}{/4.1}          & {\ul65.1}\scalebox{0.7}{/3.6}  & 77.7\scalebox{0.7}{/4.1}          & \textbf{56.7}\scalebox{0.7}{/3.4} \\
\rowcolor[HTML]{C0C0C0} 
D-CoDe (Ours)            & \textbf{80.0}\scalebox{0.7}{/4.1} & 64.2/\scalebox{0.7}{3.5}  & \textbf{79.1}\scalebox{0.7}{/4.1} & {\ul56.4}\scalebox{0.7}{/3.4}          \\ 
\bottomrule
\end{tabular}
}
\end{table}
\begin{table}[t]
\centering
\caption{Module Ablation on EgoSchema}
% {\fontsize{9.5pt}{12pt}\selectfont

{\fontsize{8pt}{9.5pt}\selectfont
% \small
\begin{tabular}{lcc}
\toprule
\multicolumn{1}{c}{Module}                                & Acc. ($\uparrow$)\\
\midrule
Baseline                                                  & 44.8             \\
\quad+ dynamic spatial token compression                  & 50.6             \\
\quad+ dynamic temporal frame selection                   & 51.8             \\
\quad+ question decomposition                             & \textbf{58.0}    \\
\bottomrule
\end{tabular}
}
\label{table:ablation}
\end{table}

\subsection{Ablation Study}

\textbf{Module Ablation.} We evaluate the contribution of each component in D-CoDe through ablation studies on the EgoSchema dataset using 15 input frames. As shown in Table~\ref{table:ablation}, all modules contribute incremental performance gains. 
The baseline adopts the naive training-free extension of LLaVA-NeXT from~\cite{llavanext-video}, which employs uniform frame sampling and spatial average pooling, yielding 44.8\% accuracy. Introducing dynamic spatial token compression increases accuracy to 50.6\% by removing redundancy while preserving salient visual cues. Incorporating dynamic temporal frame selection further improves performance to 51.8\% by prioritizing semantically diverse frames. Finally, applying question decomposition raises accuracy to 58.0\%, highlighting its effectiveness in guiding the model to focus on distinct semantic aspects of video.

\begin{table}[t]
\centering
\caption{Sampling Strategy Ablation on EgoSchema}
% {\fontsize{9.5pt}{12pt}\selectfont

{\fontsize{8pt}{9.5pt}\selectfont
% \small
\begin{tabular}{lcc}
\toprule
\multicolumn{1}{c}{Key Frame Sampling Strategy}                    & Acc. ($\uparrow$)\\
\midrule
Uniform Sampling                                         & 50.6             \\
Question-aware Sampling                               & 51.4             \\
Supplementary Frame Selection (Ours)                     & \textbf{51.8}    \\
\bottomrule
\end{tabular}
}
\label{table:sampling-ablation}
\end{table}

\begin{table}[t]
\centering
\caption{Compression Range Ablation on EgoSchema}
% {\fontsize{9.5pt}{12pt}\selectfont

{\fontsize{8pt}{9.5pt}\selectfont
% \small
\begin{tabular}{lcc}
\toprule
\multicolumn{1}{c}{Spatial Mergeable Distance Constraint}   & Acc. ($\uparrow$)\\
\midrule
$\leq 4$ neighboring tokens   & 51.4 \\
$\leq 5$ neighboring tokens   & 51.4 \\
$\leq 6$ neighboring tokens   & \textbf{52.0} \\
no constraint (Ours)          & 51.8 \\
\bottomrule
\end{tabular}
}
\label{table:compression-ablation}
\end{table}

\begin{table}[t]
\centering
\caption{Prompt Ablation on EgoSchema}
% {\fontsize{9.5pt}{12pt}\selectfont

{\fontsize{8pt}{9.5pt}\selectfont
% \small
\begin{tabular}{lcc}
\toprule
\multicolumn{1}{c}{Prompt Variant}              & Acc. ($\uparrow$)\\
\midrule
Default (Ours, as shown in Table \ref{tab:gpt3.5_prompt})   & 58.0             \\
No task/background explanation                  & 53.2             \\
Removed ``temporal and dynamic aspects''        & 54.8             \\
Rephrased (same meaning, different wording)     & \textbf{58.4}             \\
\bottomrule
\end{tabular}
}
\label{table:prompt-ablation}
\end{table}

\begin{table}[t]
\centering
\caption{Decomposed Content Ablation on EgoSchema}
% {\fontsize{9.5pt}{12pt}\selectfont

{\fontsize{8pt}{9.5pt}\selectfont
% \small
\begin{tabular}{lcc}
\toprule
\multicolumn{1}{c}{Decomposed Content}                   & Acc. ($\uparrow$)\\
\midrule
None (w/o Question Decomposition)                           & 51.8             \\
Sub-Questions                                               & 50.4             \\
Sub-Answers (Ours)                                          & \textbf{58.0}             \\
\bottomrule
\end{tabular}
}
\label{table:content-ablation}
\end{table}

\textbf{Sampling Strategy Ablation.}  
As shown in Table~\ref{table:sampling-ablation}, we evaluate frame sampling strategies on EgoSchema with D-CoDe without question decomposition. Compared to uniform sampling, question-aware sampling \cite{VaQuitA,VideoAgent} yields higher accuracy by selecting frames most semantically similar to the question using CLIP, but still performs worse than our diverse-based supplementary frame selection. This is because question-aware sampling overemphasizes query-relevant segments, limiting broader temporal modeling, whereas our method preserves more diverse visual information and thereby enhances video understanding.

\textbf{Compression Range Ablation.}  
We evaluate spatial compression range by introducing a patch distance constraint and testing on EgoSchema with D-CoDe without question decomposition. As shown in Table~\ref{table:compression-ablation}, restricting merges within $5$ patches reduces accuracy, while relaxing the constraint to $6$ patches improves performance. These results indicate that narrow merge ranges fail to compress redundant tokens effectively and hinder comprehension, whereas broader ranges reduce redundancy while preserving spatial structure, leading to improved video understanding.

\textbf{Decomposition Prompt Ablation.}  
As shown in Table~\ref{table:prompt-ablation}, we evaluate prompt variants on EgoSchema. Removing task and background explanation drops accuracy, underscoring the value of contextual guidance. Excluding ``temporal and dynamic aspects'' lowers performance, confirming the importance of temporal cues. By contrast, rephrasing with different wording yields better accuracy, suggesting that performance is determined by semantic content and is robust to structural or wording variations.

\textbf{Decomposed Content Ablation.}  
Table~\ref{table:content-ablation} evaluates the impact of decomposed content on EgoSchema. Compared with no question decomposition (None), sub-questions reduce accuracy, whereas sub-answers yield the best performance. These results indicate that the gain mainly derives from intermediate answers rather than the structured thought process, as sub-answers provide more diverse supporting content for video understanding.

\begin{figure}[t] 
    \centering
    \includegraphics[width=0.9\linewidth]{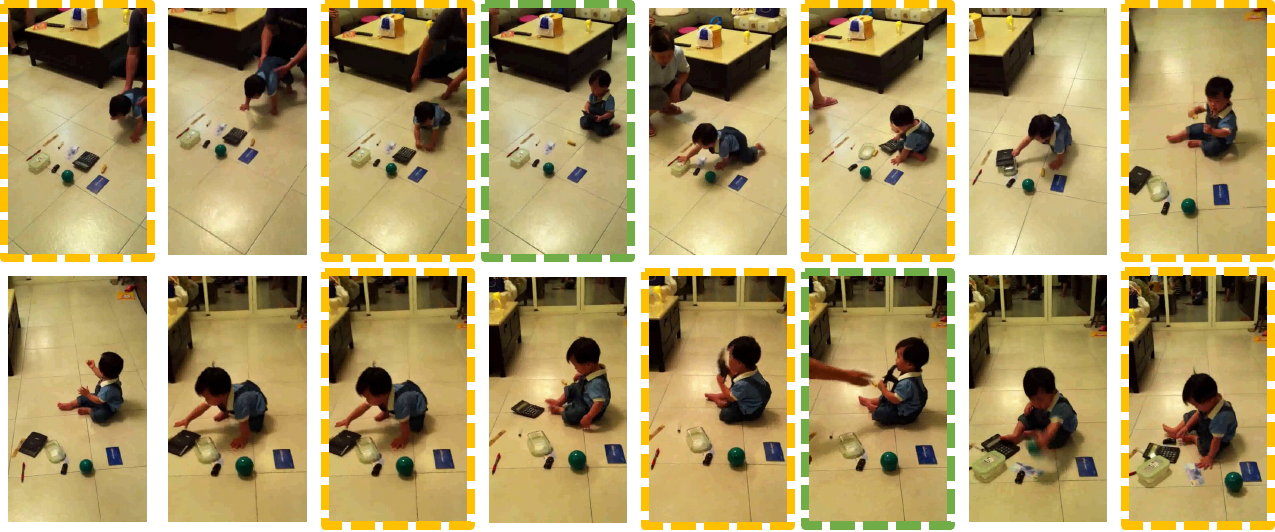}  
    \caption{Dynamic Compression (Temporal): To complement uniform sampling (yellow), supplementary frames (green) are selected from the remaining video frames based on semantic dissimilarity, thereby enhancing temporal diversity in the visual input.}
    \label{fig:compression-temporal-vis}
\end{figure}
\begin{figure}[t]
    \centering
    \resizebox{0.9\linewidth}{!}{ 

    \begin{minipage}{0.48\linewidth} 
        \begin{subfigure}{\linewidth}
            \centering
            \includegraphics[width=\linewidth]{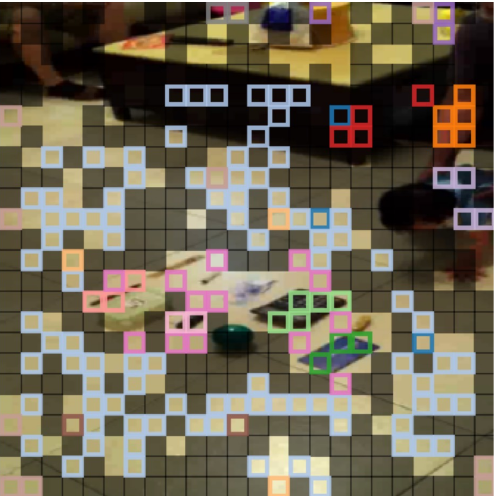}
        \end{subfigure}
    \end{minipage}
    \hfill

    \begin{minipage}{0.48\linewidth} 
        \begin{subfigure}{\linewidth}
            \centering
            \includegraphics[width=\linewidth]{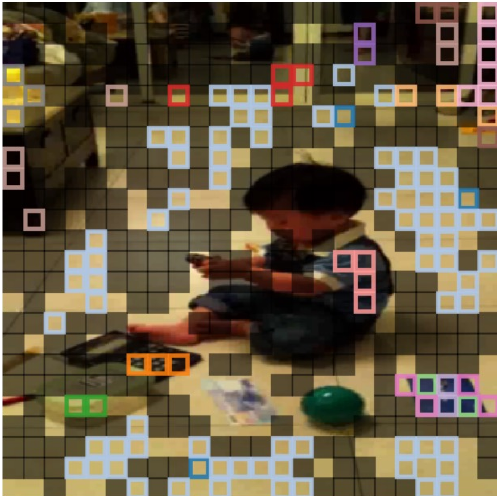}
        \end{subfigure}
    \end{minipage}
    }
    
    \caption{Dynamic Compression (Spatial): Tokens with low salience (black) are removed, and the remaining tokens are semantically clustered (indicated by color) and merged, minimizing redundancy while preserving essential visual information.}
    \label{fig:compression-spatial-vis}
\end{figure}

\begin{figure}[t]
    \centering
    \begin{subfigure}{0.9\linewidth}
        \centering
        \includegraphics[width=\linewidth]{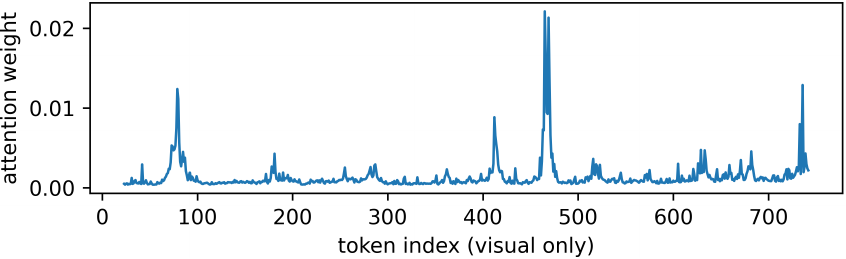}
        \captionsetup{skip=1pt} 
        \caption{Baseline w/ Original Question}
        \label{fig:att-vis-base}
    \end{subfigure}
    \vspace{1pt}

    \begin{subfigure}{0.9\linewidth}
        \centering
        \includegraphics[width=\linewidth]{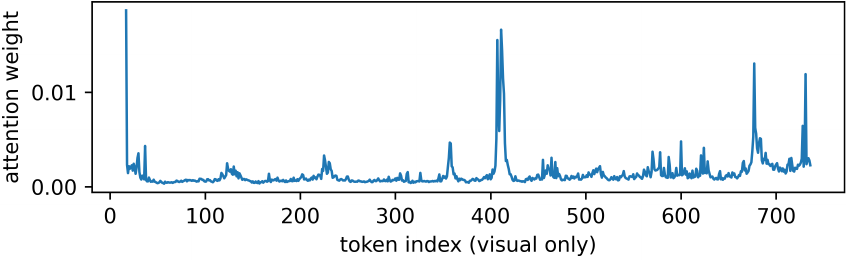}
        \captionsetup{skip=1pt} 
        \caption{Baseline w/ Sub-Question 1}
        \label{fig:att-vis-sub1}
    \end{subfigure}
    \vspace{1pt}

    \begin{subfigure}{0.9\linewidth}
        \centering
        \includegraphics[width=\linewidth]{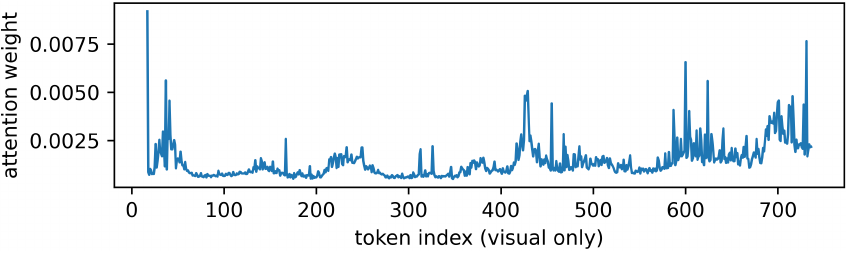}
        \captionsetup{skip=1pt} 
        \caption{Baseline w/ Sub-Question 2}
        \label{fig:att-vis-sub2}
    \end{subfigure}
    \vspace{1pt}

    \caption{Visualization of the impact of different queries on the attention distribution of the baseline model over the same visual input on IntentQA with 5 input frames. The baseline adopts the naive training-free extension of LLaVA-NeXT proposed in \cite{llavanext-video}.}
    \label{fig:att-vis}
\end{figure}

\textbf{Visualization.} 
Figure~\ref{fig:compression-temporal-vis} and Figure~\ref{fig:compression-spatial-vis} visualize the effect of D-CoDe's dynamic compression along the temporal and spatial dimensions, respectively. In the temporal dimension, supplementary frames (green), selected based on maximal semantic difference from uniformly sampled frames (yellow), enhance temporal diversity and reduce the risk of missing key actions due to fixed-interval sampling. In the spatial dimension, low-saliency tokens are discarded, while the remaining ones are clustered and merged based on semantic similarity, preserving essential content and significantly reducing redundancy. 
Figure~\ref{fig:att-vis} presents a comparison of the baseline model's attention distribution over the same visual input when prompted with the original question and its decomposed sub-questions. The observed shift in attention peaks suggests that question decomposition effectively guides the model to focus on distinct aspects of the inputs.

\section{Conclusion}  
In this paper, we investigate two key challenges in adapting image-pretrained VLMs to video understanding: the perception bottleneck, which arises from static compression strategies that uniformly process visual inputs and discard salient cues unevenly distributed across temporal and spatial dimensions; and token overload, which occurs when video inputs yield significantly more tokens than images, exceeding the model's capacity for comprehensive understanding. 
To address these challenges, we propose D-CoDe, a training-free adaptation framework that combines dynamic compression with question decomposition. Dynamic compression alleviates the perception bottleneck by adaptively selecting representative frames and performing content-aware spatial token pruning and merging, thereby preserving detail while reducing redundancy. In parallel, question decomposition mitigates token overload by reformulating complex queries into focused sub-questions that guide the model to attend to distinct aspects of the video, enabling comprehensive understanding. 
Experiments demonstrate that D-CoDe significantly improves performance on VideoQA benchmarks and shows strong potential for complex video-language tasks.

\section*{Limitations} 
The main limitation of D-CoDe lies in its relatively lower performance on videos with frequent scene transitions, compared to models employing slow-fast structures. Although D-CoDe efficiently compresses visual input and preserves key information, it still faces a trade-off between temporal and spatial retention, a limitation less evident in models such as SF-LLaVA and TS-LLaVA. To address this, future work could explore integrating a slow-fast architecture into D-CoDe to better balance temporal and spatial modeling. Additionally, incorporating a memory bank, which is commonly used in Vid-LLMs to enhance temporal awareness and maintain long-range context, may further improve the model's ability to handle complex video inputs. 

Another limitation is the difficulty in understanding durations and timestamps, a common challenge for Vid-LLMs \cite{CanMLLMTemporalUnderstanding}. While D-CoDe handles relative temporal reasoning well, precise temporal understanding remains difficult for these training-free frameworks. Addressing this may require task-specific training or architectural modifications, as shown in LLaVA-ST \cite{LLaVA-ST}.

\section*{Ethics Statement}  
The outputs of D-CoDe may occasionally contain biased or inappropriate content, potentially due to underlying biases in the base model LLaVA-NeXT \cite{llavanext}. These outputs do not reflect the authors' views. As with other generative AI systems, D-CoDe raises important ethical concerns related to content reliability and fairness. We encourage future work to implement safeguards such as dataset auditing, bias evaluation, and content attribution (e.g., watermarking), and to prioritize responsible deployment practices that balance innovation with societal impact.

\section*{Acknowledgment}  
This material is based upon work supported by the Air Force Office of Scientific Research under award number FA9550-23-1-0290.

% Bibliography entries for the entire Anthology, followed by custom entries
%\bibliography{anthology,custom}
% Custom bibliography entries only
\bibliography{custom}

\clearpage

\appendix
\section{More Background}
\subsection{Image-pretrained VLMs}
Image-pretrained vision-language models (VLMs) combine large language models (LLMs)\cite{Qwen,Learners,vicuna_report,PaLM,ChatGPT,LLaMA} with visual encoders such as CLIP\cite{clip}, enabling effective image-text alignment and multimodal understanding. As a pioneer, Flamingo~\cite{Flamingo} introduces interleaved vision-language modeling for open-ended generation. BLIP-2~\cite{blip2} employs Q-Former to align frozen vision and language models. LLaVA and its extensions~\cite{llava,llavanext} integrate lightweight connectors with instruction tuning to bridge modalities efficiently. Qwen-VL~\cite{Qwen-VL,Qwen2-VL,Qwen25-VL} connects a visual encoder to the Qwen language model via cross-attention, supporting high-resolution and multilingual reasoning. Most recently, MM1~\cite{MM1} provides a systematic analysis of model scaling and data design for efficient VLM training.
\begin{table}[t]
\centering

{\fontsize{8pt}{10pt}\selectfont
\begin{minipage}{0.48\linewidth}
    \centering
    \caption{Ablation of $\alpha$}
    \label{tab:alpha-ablation}
    \begin{tabular}{c|c}
        \toprule
        $\alpha$ & Acc. ($\uparrow$) \\
        \midrule
        0.80 & 57.2 \\
        0.85 & \textbf{58.0} \\
        0.90 & 57.0 \\
        0.95 & 56.2 \\
        \bottomrule
    \end{tabular}
\end{minipage}
}
\hfill
{\fontsize{8pt}{10pt}\selectfont
\begin{minipage}{0.48\linewidth}
    \centering
    \caption{Ablation of $\beta$}
    \label{tab:beta-ablation}
    \begin{tabular}{c|c}
        \toprule
        $\beta$ & Acc. ($\uparrow$) \\
        \midrule
        0.575 & 56.6 \\
        0.600 & 56.4 \\
        0.625 & \textbf{58.0} \\
        0.650 & 56.2 \\
        \bottomrule
    \end{tabular}
\end{minipage}
}
\vspace{0.5em}

{\fontsize{8pt}{10pt}\selectfont
\begin{minipage}{0.48\linewidth}
    \centering
    \caption{Ablation of $\tau$}
    \label{tab:tau-ablation}
    \begin{tabular}{c|c}
        \toprule
        $\tau$ & Acc. ($\uparrow$) \\
        \midrule
        0.80 & 55.0 \\
        0.85 & 55.6 \\
        0.90 & \textbf{58.0} \\
        0.95 & 57.0 \\
        \bottomrule
    \end{tabular}
\end{minipage}
}
\hfill
{\fontsize{8pt}{10pt}\selectfont
\begin{minipage}{0.48\linewidth}
    \centering
    \caption{Ablation of $t$}
    \label{tab:t-ablation}
    \begin{tabular}{c|c}
        \toprule
        $t$ & Acc. ($\uparrow$) \\
        \midrule
        0.3 & 56.0 \\
        0.5 & \textbf{58.0} \\
        0.7 & 56.2 \\
        0.9 & 56.4 \\
        \bottomrule
    \end{tabular}
\end{minipage}
}
\end{table}

\section{Hyper-Parameters Ablation}

\subsection{Effect of $\alpha$ in Dynamic Compression}
Table~\ref{tab:alpha-ablation} presents the results of an ablation study on $\alpha$, which controls the uniform sampling ratio in temporal dynamic compression. The results show that accuracy on the EgoSchema benchmark increases significantly as $\alpha$ decreases from 0.95 to 0.85. This suggests that selecting supplementary key frames to emphasize informative segments enhances the video understanding ability of image-pretrained VLMs. However, when $\alpha$ continues to decrease, the number of uniformly sampled frames becomes insufficient, weakening the model's global perception of the video and leading to performance degradation.

\begin{table*}[t]
\centering
\caption{Module Ablation on Other Benchmarks.}
{\fontsize{8pt}{10pt}\selectfont
\begin{tabular}{l cc cccc}
\toprule
\multirow{2}{*}{Module} & \multicolumn{2}{c}{Multi-choice VideoQA (Acc., $\uparrow$)} & \multicolumn{4}{c}{Open-ended VideoQA (Acc./Score, $\uparrow$)} \\ \cmidrule(lr){2-3} \cmidrule(lr){4-7}
& NExT-QA & IntentQA & MSVD & MSRVTT & TGIF & ANet \\
\midrule
Baseline                                & 65.4 & 61.3 & 77.8/\scalebox{0.7}{4.0} & 62.8/\scalebox{0.7}{3.5} & 76.9/\scalebox{0.7}{4.0} & 54.2/\scalebox{0.7}{3.3} \\
\quad + Dynamic Spatial Token Compression & 66.7 & 62.2 & 79.4/\scalebox{0.7}{4.0} & 63.6/\scalebox{0.7}{3.5} & 78.9/\scalebox{0.7}{4.1} & 55.4/\scalebox{0.7}{3.3} \\
\quad + Dynamic Temporal Frame Selection  & 67.0 & 62.9 & \textbf{80.0}/\scalebox{0.7}{4.1} & \textbf{64.2}/\scalebox{0.7}{3.5} & \textbf{79.1}/\scalebox{0.7}{4.1} & \textbf{56.4}/\scalebox{0.7}{3.4} \\
\quad + Question Decomposition            & \textbf{68.3} & \textbf{64.2} & 72.4/\scalebox{0.7}{3.8} & 62.2/\scalebox{0.7}{3.5} & 75.7/\scalebox{0.7}{4.0} & 53.8/\scalebox{0.7}{3.3} \\
\bottomrule
\end{tabular}
}
\label{table:ablation-all}
\end{table*}

\subsection{Effect of $\beta$ in Dynamic Compression}
Table~\ref{tab:beta-ablation} presents the results of an ablation study on $\beta$, which controls the proportion of spatial tokens retained during pruning. A larger $\beta$ corresponds to more aggressive pruning, keeping only highly activated tokens. As $\beta$ increases, the accuracy on the EgoSchema benchmark first rises and then falls, indicating that retaining too few tokens harms the model's understanding of visual content, while retaining too many introduces redundant noise.

\subsection{Effect of $\tau$ in Dynamic Compression}
Table~\ref{tab:tau-ablation} presents the results of an ablation study on $\tau$, the cosine similarity threshold used for token merging. A smaller $\tau$ results in more tokens being grouped and merged. As $\tau$ increases, accuracy on the EgoSchema benchmark follows a rise-then-fall trend. This is because merging low-similarity tokens can blur critical visual details, while merging only highly similar tokens leads to token redundancy, both of which hinder model performance.

\subsection{Effect of $t$ in Question Decomposition}
Table~\ref{tab:t-ablation} presents the results of an ablation study on $t$, which controls the diversity of sub-questions generated during question decomposition. A larger $t$ results in more diverse sub-questions. Accuracy on the EgoSchema benchmark initially improves as $t$ increases, but eventually declines. This indicates that both insufficient and excessive diversity can impair the model's ability to comprehensively interpret large volumes of visual tokens.

\begin{table}[t]
\centering
% \small
{\fontsize{8pt}{10pt}\selectfont
\caption{Question Decomposition Example on MSVD}
\begin{tabular}{p{7.25cm}}
\toprule
\textbf{Original Question:} \\
"What is a man sitting on?" \\ \midrule

\textbf{Decomposed Sub-Questions:} \\ 
1. Does the man change location during the video, or does he remain in one place?  \\
2. At what point does he begin sitting down, and what happens before that?  \\
3. What object does he touch when sitting down?  \\
4. Does that object stay consistent across the video? \\ 
5. Do we observe interactions confirming it’s a seat? \\ 
6. Do any video perspectives help reveal more detail? \\
\bottomrule
\end{tabular}
\label{tab:sub-ques-example}
}
\end{table}
\section{Additional Module Ablation}
Table~\ref{table:ablation-all} reports module-wise ablations on additional benchmarks, including NExT-QA, IntentQA, MSVD, MSRVTT, TGIF, and ANet. For multi-choice VideoQA, all modules yield incremental gains. For open-ended VideoQA, where questions are generally simpler (Section~\ref{sec:open-end-results}), question decomposition can mislead the model.  

Table~\ref{tab:sub-ques-example} shows an MSVD case where decomposition, though semantically valid, overcomplicates a simple spatial query and lowers accuracy. This indicates that decomposition is most effective for complex or multi-step reasoning.

\begin{table}[t]
\centering
\caption{Error Analysis on MSRVTT-QA}
{\fontsize{8pt}{10pt}\selectfont
\begin{tabular}{lcc}
\toprule
\multicolumn{1}{c}{Method} & Full Set & Top-100 Scene Change Samples \\
\midrule
SF-LLaVA & 65.8/\scalebox{0.7}{3.6} & 64.0/\scalebox{0.7}{3.5} \\
D-CoDe (Ours)   & 64.2/\scalebox{0.7}{3.5} & 56.0/\scalebox{0.7}{3.3} \\
\bottomrule
\end{tabular}
}
\label{table:msrvtt-error-analysis}
\end{table}

\section{Error Analysis}
As noted in Section~\ref{sec:open-end-results}, D-CoDe performs worse on videos with frequent scene transitions. Table~\ref{table:msrvtt-error-analysis} shows results on MSRVTT-QA using the full set and 100 samples with the most transitions. SF-LLaVA remains stable (65.8 vs. 64.0), whereas D-CoDe drops sharply (64.2 vs. 56.0), confirming its sensitivity to rapid scene changes.

\begin{table}[t]
\centering
\caption{Efficiency Analysis on EgoSchema}
{\fontsize{8pt}{10pt}\selectfont
\begin{tabular}{lccc}
\toprule
Module & Acc. ($\uparrow$) & s/sample ($\downarrow$)\\
\midrule
Baseline & 44.8 & 3.927 \\
+ Dynamic Compression & 51.8 & 6.115  \\
+ Question Decomposition & 58.0 & 37.395  \\
\bottomrule
\end{tabular}
}
\label{tab:latency-analysis}
\end{table}
\begin{table}[t]
\centering
\caption{Trade-off Analysis on EgoSchema}
{\fontsize{8pt}{10pt}\selectfont
\begin{tabular}{lccc}
\toprule
Module & Acc. ($\uparrow$) & s/sample ($\downarrow$)\\
\midrule
D-CoDe & 58.0 & 37.395  \\
w/ smaller CLIP (35\% params) & 58.2 & 35.466 \\
w/ Limit sub-question count = 5 & 56.0 & 26.273  \\
w/ Limit sub-question count = 7 & 57.8 & 33.704  \\
\bottomrule
\end{tabular}
}
\label{tab:light-dcode}
\end{table}
\section{Efficiency Analysis}
Table~\ref{tab:latency-analysis} shows latency and accuracy results on EgoSchema. Compared with the baseline, dynamic compression increases inference time slightly, whereas question decomposition causes a larger latency increase.  

Although D-CoDe introduces additional inference overhead, the performance–cost trade-off can be adjusted through simple design choices. Table~\ref{tab:light-dcode} further evaluates lighter variants, showing that using a lightweight visual encoder (CLIP-ViT-B/32, 35\% parameters) for supplementary frame selection or restricting the number of sub-questions reduces inference time substantially while maintaining competitive accuracy.

\begin{table}[!t]
\centering
% \small
{\fontsize{8pt}{9pt}\selectfont
\caption{Prompt Variant used in Table \ref{table:prompt-ablation}}
\begin{tabular}{p{7.25cm}}
\toprule
\textbf{No task/background explanation:} \\
Your job is to break down the given question into a series of subquestions that guide the model toward solving the problem. The subquestions should focus on temporal and dynamic aspects of the video, rather than just static information. \\
\\
Question: ``\{\textit{user question here}\}'' \\
\\
Output the subquestions as a Python list of strings. \\ \midrule

\textbf{Removed ``temporal and dynamic aspects'':} \\
I am working on a video understanding task. Your job is to break down the given question into subquestions that guide the model toward solving the problem. I will provide a question, and you should output the corresponding subquestions in English. \\
\\
Question: ``\{\textit{user question here}\}'' \\
\\
Output the subquestions as a Python list of strings. \\ \midrule

\textbf{Rephrased:} \\ 
Your task is to break down the given video understanding question into a series of subquestions. These subquestions are crucial for guiding the model and **must prioritize temporal and dynamic aspects** of the video. Crucially, they should **not rely on static information** obtainable from a single frame. \\
\\
Output the subquestions as a **Python list of strings**. Each subquestion should focus on the **evolution, changes, and interactions over time** within the video. \\
\\
Question: ``\{\textit{user question here}\}'' \\
\\
Output: \\

\bottomrule
\end{tabular}
\label{tab:prompt-variant}
}
\end{table}
\section{Detailed Prompt Variants}
Table~\ref{tab:prompt-variant} lists the prompt variants used in the Decomposition Prompt Ablation (Table~\ref{table:prompt-ablation}).
\end{document}